# Minimal Pair-Based Evaluation of Code-Switching

**Igor Sterner*** and **Simone Teufel**
Department of Computer Science and Technology
University of Cambridge
United Kingdom
{is473,sht25}@cam.ac.uk

## Abstract

There is a lack of an evaluation methodology that estimates the extent to which large language models (LLMs) use code-switching (CS) in the same way as bilinguals. Existing methods do not have wide language coverage, fail to account for the diverse range of CS phenomena, or do not scale. We propose an intervention based on minimal pairs of CS. Each minimal pair contains one naturally occurring CS sentence and one minimally manipulated variant. We collect up to 1,000 such pairs each for 11 language pairs. Our human experiments show that, for every language pair, bilinguals consistently prefer the naturally occurring CS sentence. Meanwhile our experiments with current LLMs show that the larger the model, the more consistently it assigns higher probability to the naturally occurring CS sentence than to the variant. In accordance with theoretical claims, the largest probability differences arise in those pairs where the manipulated material consisted of closed-class words.[1]

## 1 Introduction

Bilinguals take advantage of knowledge of all the languages available to them in a given situation. The phenomenon of code-switching (henceforth CS) occurs when they choose to use more than one language in the same sentence. A particularly interesting form of CS involves syntactic dependencies between material of different languages, rather than just a simple insertion of, for instance, a technical term in one language into the other. In the literature, this type of CS is called alternation or congruent lexicalization, as opposed to insertion (Muysken, 1997).

CS offers bilinguals substantial communicative advantages, such as the ability to express oneself more precisely, more quickly or even with added humour (Gumperz, 1982; Myers-Scotton, 1993). In many communities, it is even the predominant form of communication (Grosjean, 2010). As a result, many users will want to instruct LLMs and get responses from LLMs using CS that they deem natural. LLMs can of course already generate mixed-language text, as in the experiments by Yong et al. (2023) and Kuwanto et al. (2024). But the point is not whether a text produced by any system contains words from more than one language – it is rather whether they are mixed in the way that a bilingual human might.

Previous approaches to CS evaluation exist, via specially-created sets of human judgments (e.g., Yong et al., 2023; Kuwanto et al., 2024; Kodali et al., 2024). In these approaches, participants were asked to score how acceptable each sentence appears to them, using a numeric score. However, it is well-known that the assignment of absolute scores is highly subjective and influenced by a wide range of other factors (Schütze, 1996; Keller and Asudeh, 2002). Therefore, such scores are not comparable across sentences, because the participant has not seen these sentences in comparison. Given the known context sensitivity and thus subjectivity of CS, this substantially lowers the value of the numerical scores for quantifying the quality of CS text. Inter-annotator agreement is also unsatisfactorily low (Kuwanto et al., 2024).

Kuwanto et al. (2024) aim to generate more natural-sounding CS text by LLM prompting, guided by theoretical constraints according to Poplack's theory (1980). This theory predicts points in the sentence where the language is likely to change. They then use prompting to create variants of a monolingual sentence using these switch points. Each generated sentence is judged independently by humans, and later $O(n \times m^2)$ pairs of CS sentences ($n$ is the number of original monolingual sentences and $m$ is the number of LLMs used) are created. In principle, the availability of several

---
*The author is now at the University of Edinburgh.
[1]Benchmark, corpora and code are available from https://github.com/igorsterner/acs.

(1) a. @USER **And I said maybe** <u>etwas</u> leiser singen, sonst ruf ich die Polizei
    b. @USER **And I said maybe** <u>a little</u> leiser singen, sonst ruf ich die Polizei

[German–English, a. from Sterner and Teufel, 2023]

(2) a. **Would do it** <u>myself</u> om inte make var bilmek och nördig med det.
    b. **Would do it** <u>själv</u> om inte make var bilmek och nördig med det.

[Swedish–English, a. from DeLucia et al., 2022]

(3) a. 同学们 会 大概 <u>知道</u> **what they want to do in the future**
    b. 同学们 会 大概 <u>know</u> **what they want to do in the future**

[Chinese–English, a. from Lyu et al., 2010]

CS variants and the idea of pairing the generations might alleviate the problem of non-comparability. However, the methodological problem of absolute scoring remains. The participant, as before, operated with absolute scores on individual sentences, never seeing the two sentences in a pair together and comparing them. Additionally, within each pair, the number of differences is not constant, due to the unpredictability of the LLM. This makes it almost impossible to assign the naturalness of any particular generation to syntactic or lexical changes the LLM made to that sentence.

Meanwhile in the theoretical community working on the acceptability of monolingual text, minimal pairs are the accepted approach for making controlled comparisons (Schütze, 1996; Keller and Asudeh, 2002). Recently this was applied in the BLiMP benchmark (Warstadt et al., 2020), which provided a reliable way to track the progress of LLMs. It is now used to track the progress of smaller LLMs trained on volumes of data more comparable to what is required for human language acquisition (Warstadt et al., 2023; Hu et al., 2024; Charpentier et al., 2025). Similar benchmarks have also since been proposed in other languages (Xiang et al., 2021; Volodina et al., 2021; Song et al., 2022; Jentoft and Samuel, 2023; Taktasheva et al., 2024; Suijkerbuijk et al., 2025; Liu et al., 2024; Jumelet et al., 2025).

An important principle of the minimal-pair methodology is that the difference in each pair is carefully controlled: the two sentences are identical except for one single difference. Therefore, if there is a difference in acceptability between the sentences, it can be attributed directly to the single textual difference. In the current paper, we will define CS acceptability as the difference between sentences in minimal pairs of CS. However it is not currently possible to theoretically define acceptability in CS as precisely as for monolingual settings. Judgments of CS acceptability are more subjective than those of monolingual grammaticality, and we do not have a theoretical reference for what bilinguals tend to judge as acceptable, although they themselves have strong intuitions in individual sentences as to what CS they deem sounds natural (Joshi, 1982).

Another problem is scalability. To fairly assess how a system models the near-infinite number of linguistic variations possible, as many fundamentally different pairs in the evaluation as possible are needed. It is therefore best to have only one pair for each original sentence, unlike in the approach of Kuwanto et al. (2024).

We propose that the solution to the problem is to algorithmically create negative samples derived from naturally occurring CS, forming minimal pairs. This allows us to create as many minimal pairs as we have suitable CS sentences. We can then use the expensive human judgments more effectively, namely by ensuring that each judgment is based on a CS sentence pair that is fundamentally independent of all other pairs in that experiment. Once human judgments have been collected to validate our method of creating minimal pairs, further judgments are no longer required; human judgments are a one-time effort.

In this paper, we

1. Create a benchmark corpus of minimal pairs of CS covering 11 language pairs. Three examples are given in Examples 1-3 (for glosses and translations see Appx. C.1); 1a, 2a and 3a are sampled from naturally occurring, observed CS, whereas 1b, 2b and 3b are algorithmically-derived variants (bold words are English).

2. Validate it with human judgments, showing that bilinguals consistently prefer the observed CS in each minimal pair. We also show that bilinguals agree with one another on the task, to a reasonable degree ($\kappa$=0.57).

3. Evaluate current LLMs on the benchmark. Most LLMs we test perform poorly. The best-performing one was the largest LLM we run (`Llama-3.1 405B`). In accordance with theoretical claims, our analysis also shows a consistent trend that minimal pairs involving manipulating closed-class words leads to larger differences in LLM probability, as compared to open-class words.

## 2 Minimal Pairs of CS

We assume that all observed CS is acceptable, and we can get a near-unlimited supply of these from social media. We propose to manipulate observed CS sentences, under the assumption that on average this manipulation will decrease acceptability. Each naturally observed CS sentence and a manipulated variant of that sentence together form a minimal pair.

Our method should (1) result in minimal pairs with noticeably different acceptability but where both sentences are syntactically plausible (2) be language-agnostic, and (3) generate a syntactically and semantically diverse set of minimal pairs. We also need to consider sentence complexity. An example of such complexity difference would be if the number of times the language switches is not the same in the two sentences. Condition (4) is therefore that the sentence complexity inside a pair should be comparable.

We collect a large pool of suitable CS sentences where the two languages are grammatically integrated. We start from tweets which have been identified by Twitter as non-English and identify those which nevertheless contain English words. We will consider each word to have exactly one *CS status*. Either it belongs to either one or the other of the two languages concerned, or if its status is unclear (words of mixed morphology, named entities, punctuation), its label should be 'Other'. Since our goal is to collect intrasentential CS, each example should cover exactly one sentence. A sentence segmentation tool is necessary at this stage, as social media posts often contain more than one sentence.

We next identify the *switch point* in the sentences.[2] Our manipulation consists in translating one single word adjacent to this switch point into the other language. This operation can be carried out for any language pair, provided a translation is available. During the process, we only generate minimal pairs that have the same number of switch points within the sentence pair, as this keeps sentence complexity constant.

### 2.1 Data Collection

For the language pair German–English, we start from the most recent 27.5 million raw tweets used in our earlier corpus (Sterner and Teufel, 2023). We also treat other languages code-switched with English, using the raw Twitter data collected by DeLucia et al. (2022). The languages we treat must be written in Latin script and there must exist at least 100K raw tweets. This results in data for Danish, Spanish, French, Indonesian, Italian, Dutch, Swedish and Turkish. Geographically, these languages are all European (except 1) and genetically they are all Indo-European (except 1). We include up to 10 million tweets for each language.

Token-level language identification (LID) is provided by our `AnE` system (Sterner, 2024), which represents the SoTA for this task on CS text. This step labels every word with a label expressing its CS status: 'Lang1' (the respective non-English language), 'English', 'Mixed', 'Named Entity', and 'Other'; the final three are collated in the 'Other' category. We now identify borrowed words using lists from Wiktionary (Ylonen, 2022)[3], and remove tweets containing obscene words using fixed lists. We also perform some normalisation to the tweets, including converting one or more consecutive Twitter mentions to a single '@USER'. Emojis are kept, as are hashtags.

We then apply a second, more refined, LID step. This became necessary as we found that many of the resulting tweets did not match the language assigned to them by Twitter's LID algorithm. We process the string of all 'Lang1' words in a sentence (with other words removed) with Lingua[4], the SoTA for identifying the language of monolingual short sequences of words. Tweets where Lingua's language prediction differs from Twitter's original prediction are removed. We then sentence-segment the tweets using our existing neural sentence segmentation model (Frohmann et al., 2024), specifically, the `sat-12-sm` model. This segmentation model is the first that has been shown to work satisfactorily on CS text. Frohmann et al. evaluated the model on four CS corpora, with all input text lowercased and punctuation removed. It achieved a macro average F1 score (on sentence-ending tokens) of 54.4%, as compared to prompting `Llama-3.1 8B` which achieved 43.4% or using Spacy's dependency parsing-based approach which achieved only 12.2%. Resulting sentences that are greater than 200 characters or fewer than 6 tokens are removed.

---
[2]Switch points in the examples therefore correspond to the change between normal and bold font.
[3]See Appendix Table 3 for statistics.
[4]https://github.com/pemistahl/lingua-py

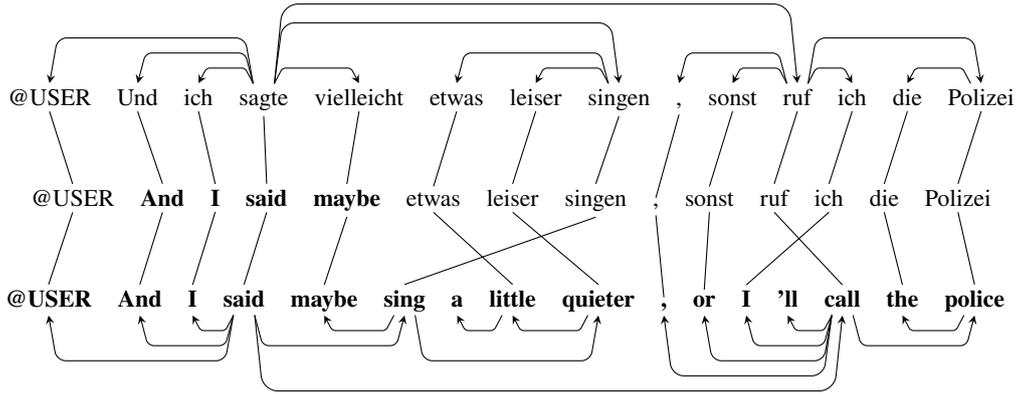

Figure 1: An example CS sentence during automatic processing. In this case we generate minimal pair 1.

Our choice of languages was due to the fact that the `AnE` tool only operates on language pairs where one language is English, and where the other language is also in Latin script. To represent more languages, we complement our corpus with two further language pairs, Turkish–German (data and goldstandard CS status labels from Çetinoğlu (2016)) and Chinese–English (data from Lyu et al. (2010) and Lovenia et al. (2022)). Both originate from transcriptions of spoken language. The domain of spoken CS language has both similarities with and differences from the domain of written CS (Gardner-Chloros and Weston, 2015); including these two corpora will enable us to later investigate this further.

We chose Turkish–German because we are already treating German–English and Turkish–English CS, and so it will be interesting to compare these three language pairs. Chinese–English is relevant because this pair concerns languages that are typologically far-removed from each other. For the Chinese–English sentences, we determine word-level language and CS status with a regular expression by marking all Chinese characters as Lang1. More details of the data collection can be found in Appendix A.

## 2.2 Generating Minimal Pairs

As a first step, the CS sentences are translated into monolingual sentences of each of the contributing languages. Most translation systems require the specification of a single input language. With CS input, this is not possible. The `madlad-3B` machine translation (MT) system from Kudugunta et al. (2023) does not take any language specification as input and is near-SoTA in open-source MT, so we use this model. We also require word-to-word alignments between the CS sentence and the monolingual sentences. This is accomplished by the `awesome-align` model (Dou and Neubig, 2021).

We next perform named-entity recognition (NER) and dependency parsing on the monolingual sentences with Qi et al.'s tools (2020). If a token in the CS sentence is aligned to a named entity in either of the monolingual sentences, we change its CS status to 'Other', unless `AnE` already marked it as 'Other'. Multi-word expressions are tagged using a list from the Urban Dictionary. Figure 1 shows sentence 1a (p. 2) during this stage of processing.

On the basis of the dependency parse, we remove non-integrative CS sentences, i.e., those where there is no syntactic link between the less frequent material and the rest of the sentence. For the remainder, we use the alignment to create minimal pairs.

The alignment can be one-to-many or many-to-one. For instance, due to different compounding morphologies across languages, a situation can arise where a single compound word in one language is translated into many words in the other language. Furthermore, alignment errors can sometimes result in unaligned words in a translation. In the rare cases when such unaligned material happens to appear next to a word that we have decided to move the switch point over, we move the switch point over the unaligned material and its neighbour (i.e. include the unaligned material), but only if there is a syntactic dependency between this word and the neighbour. This results in syntactically more plausible minimal pairs.

In the example in Figure 1, the switch point occurs between the English adverb 'maybe' and the German adverb 'etwas'. The German adverb is aligned with its translation 'little', and its neighbour 'a' is not aligned. The syntactic dependency between 'a' and 'little' licenses us to move the switch point over both, yielding the sentence in Example 1b.

We also decided to disallow switches where there is a noun on either side of the switch point. This is because such cases often lead to problems with keeping CS complexity comparable. While this is not ideal, we feel it is better than the alternative, which would be to sample minimal pairs uniformly across POS. There is a diverse range of observed CS phenomena which we try to capture with our benchmark (see examples in Appendix C). As this range is not naturally uniform over POS, we decided against POS-based sampling but excluded nouns.

With our approach, there may be more than one possible minimal pair for each input CS sentence. Furthermore, of all possible minimal pairs, certain replacements will be more frequent than others. To mitigate the effect of this, we enforce that each individual lexical difference occurs only once for each language pair.

For evaluation, we sample up to 1,000 minimal pairs each for the 11 language pairs, and call the result the ACS (Acceptability of Code-Switching) benchmark. Corpus statistics and examples for each language pair are given in Appendix C.

We aimed for the highest quality of minimal pairs in each language pair. However, the available tools necessary for creating minimal pairs do not perform equally across all language pairs. The tools are validated to perform best in German–English and Spanish–English. This is because of the availability of CS training data for these language pairs, and the general high-resource status of all three contributing languages.

We take steps with the aim of removing poorly processed data. One case concerns sentences incorrectly identified to contain any CS. We remove cases where one of the monolingual translations has a Levenshtein distance of fewer than 5 characters with the source CS sentence. This also removes some cases where there is CS remaining after the translation. Based on the POS-analysis of the monolingual translations, we also remove cases with at least one word with the 'X' tag. Such a tag indicates the presence of unintelligible, fragmented or untranslated material.

We performed an error analysis on the automatic tools used. For this, the second author of this paper analysed 100 German–English minimal pairs and the output of the tools.[5]

The analysis assessed for each minimal pair if there was an error in the processing that would make it obvious which sentence in the pair was the manipulated sentence. If this was not the case, the pair was labelled as "fair". If there was an error, the annotator determined the first tool in the pipeline that caused the error (order: sentence segmentation, tokenization, language identification, translation, alignment, minimal pair creation algorithm).

Out of 100 sentences, 84 were genuine CS sentences where the meaning was clear; in 9 cases, the sentence was monolingual, and in 7, the meaning of the sentence was unrecoverably unclear, mostly due to typos in the input. In 76% of genuine CS sentences, the pair was determined to be fair; in 70%, all tools in the pipeline worked error-free. Errors from the tools were most frequent with the automatic translations (10) and alignment (5). Segmentation (2), tokenization (1) and language identification (1) were less of a problem. In 6 cases, the data generation procedure was the problem, as it did not correctly handle complex German or English morphology or subcategorisation.

### 2.3 Human Data Validation

The definition of truth in our benchmark comes from the fact that one of the sentences has naturally occurred in a CS discourse. We now measure agreement of one human participant per language pair with this gold standard. As a secondary measurement, for German–English, we also provide the agreement of 3 human participants amongst each other. Since collecting judgments from many participants is expensive, we measure inter-annotator

---

[5]A visualisation similar to Figure 1 was provided.

|   | 1 | 2 | 3 | 4 | 5 |
|---|---|---|---|---|---|
| $\kappa$ | 0.58 | 0.69 | 0.58 | 0.52 | 0.53 |
| Acc (%) | 80.6 | 84.6 | 79.1 | 76.1 | 76.6 |

Table 1: Human agreement with the gold standard for German–English

|   | da-en | es-en | fr-en | id-en | it-en |
|---|---|---|---|---|---|
| $\kappa$ | 0.66 | 0.70 | 0.50 | 0.72 | 0.73 |
| Acc (%) | 83.1 | 85.1 | 75.1 | 86.1 | 86.6 |
|   | nl-en | sv-en | tr-en | tr-de | zh-en |
| $\kappa$ | 0.62 | 0.64 | 0.71 | 0.75 | 0.59 |
| Acc (%) | 81.1 | 82.1 | 85.6 | 87.6 | 79.6 |

Table 2: Human agreements with the gold standard for the 10 other language pairs

agreement only for this language pair.

We needed bilingual or near-bilingual participants who were likely to employ CS in their own daily lives. We therefore recruited individuals whose mother tongue is not English and who have lived in the US or UK for at least one year at one point in their lives. All were university students at master's or PhD level; twelve were between 18 and 25, three between 26 and 35 years old.

For each language pair, a sample of the respective ACS benchmark was used; the place inside the minimal pair where the original sentence appeared was randomised. In the languages other than German–English, where single annotation took place, each participant worked on 201 (67*3) minimal pairs. For German–English, minimal pairs are drawn uniformly from a pool of 5 participants, such that each of 335 pairs is annotated by three unique participants. Each participant thus also worked on 201 (67*3) minimal pairs. The difference between the two sentences in a minimal pair was bold-faced in order to facilitate the decision. Annotators worked in batches of 67 minimal pairs in their own time. The acceptability judgment task was operationalised as follows: We told annotators to select from a sentence pair the one that had been produced by another bilingual. We also told them that the other sentence had been altered.

We use accuracy to compare human annotation against the gold standard. Accuracy is the proportion of minimal pairs for which the participant selected the observed CS sentence. We also use Fleiss's kappa (1971).

For German–English, inter-annotator agreement was measured at $\kappa = 0.57$ (N=335, n=2, k=3). This is well within the range of what is typical in the field (Kuwanto et al.: Krippendorff's $\alpha$=0.32 (Tamil-English), $\alpha$=0.41 (Malayalam-English), $\alpha$=0.65 (Hindi-English)). The judgments were drawn from a pool of 5 participants who each annotated 201 minimal pairs. Table 1 gives individual results. Accuracy ranges between 76.1-84.6% and $\kappa$ between 0.52-0.69. In 205 out of the 335 cases, all three annotators agreed with the gold standard. For 22, all 3 annotators chose the manipulated sentence (listed in Appx. C.13). At least one annotator agreed with our gold standard 93.4% of the time.

Table 2 shows results of the human experiments between the gold standard and the human participants on the other language pairs. We find reasonable agreement, with $\kappa > 0.6$ for 8/10 language pairs and $\kappa > 0.5$ for the remaining two.

The results confirm our expectations that bilinguals across many language pairs have shared knowledge of what CS other bilinguals would produce. Their intuitions also often align with the gold standard. Our method of generating minimal pairs had the goal of being cross-lingually applicable, and our human judgments bring supporting evidence for this claim. Although Turkish belongs to a different language family than English or German, we observe $\kappa > 0.7$ and accuracies of 85.6% and 87.6% for Turkish–English and Turkish–German.

## 3 Experiments on Automatic Evaluation

We demonstrate our evaluation methodology with a comparison of several open-weight LLMs.

### 3.1 Setup

We selected five families of multilingual and open-weight autoregressive LLMs for our investigation: `BLOOM` (BigScience, 2023), `Llama-3` (MetaAI, 2024), `Qwen2.5` (Qwen, 2025), `OLMo` (AI2, 2024) and `EuroLLM` (Martins et al., 2024). Given its high performance in monolingual benchmarks and Kodali et al. (2024) finding it works best for CS, we also compare against `XLM-RoBERTa` multilingual masked language models (Conneau et al., 2020). For each family, we evaluate base models with full precision[6] for all the model sizes available.

Our metric is accuracy, which is the proportion of pairs, (consisting an observed CS sentence $s^o$ and a manipulated variant $s^m$, from the challenge set of a language pair, $\mathcal{S}$), for which the LLM, $\mathcal{M}$, assigns higher probability to the observed CS sentence:

$$\text{Acc}(\mathcal{M}; \mathcal{S}) = \frac{1}{|\mathcal{S}|} \sum_{s^{o,m} \in \mathcal{S}} \mathbb{1}\left[P_\mathcal{M}(s^o) > P_\mathcal{M}(s^m)\right]$$

---
[6] Except the 405B version of `Llama-3.1`, for which we use a version with all weights stored as normalized 4-bit floats (Dettmers et al., 2023). All runs were performed on one node with up to 4 A100-80GB GPUs.

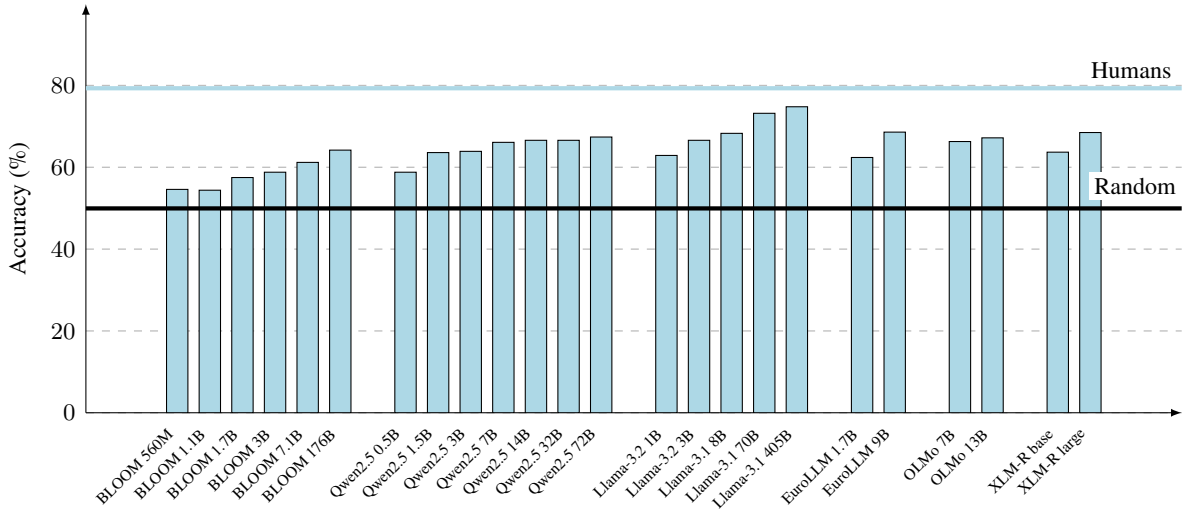

Figure 2: ACS (de-en) model performance.

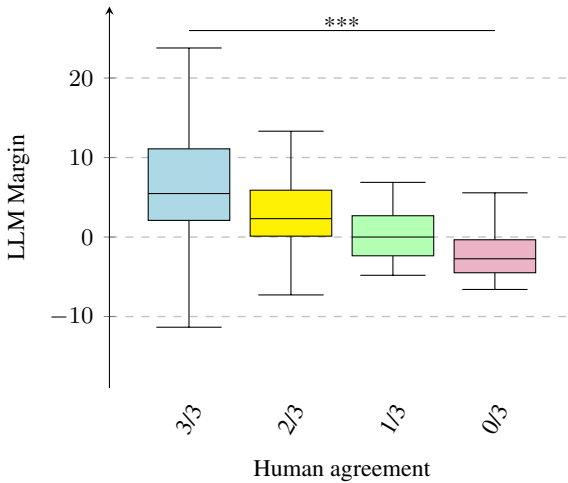

Figure 3: Human agreement with the gold standard vs. LLM margin (de-en).

Autoregressive LLMs model sentence probability directly. For the masked language models, we used the pseudo-log-likelihood formulation from Kauf and Ivanova (2023).[7] We also define the margin of an LLM prediction on a minimal pair as the difference in log probability between the sentences in the minimal pair:

$$\text{Margin}(\mathcal{M}; s^{o,m}) = \log P_\mathcal{M}(s^o) - \log P_\mathcal{M}(s^m)$$

We test differences in system accuracy for significance using the paired two-tailed Monte-Carlo permutation test with $R = 10,000$ and $\alpha = 0.05$. We will test differences between groups of LLM margins and differences between human and system judgments for significance using an unpaired version of the same permutation test.

### 3.2 Results

**LLM Accuracy** Figure 2 gives results for German–English. The largest `Llama` model, 3.1 405B, is significantly better than all its smaller counterparts (A=74.8%, compared to A=62.9%, A=66.6%, A=68.3% and A=73.2%, all with p<0.05); it is also better than all other tested LLMs (all with p<0.001). However, performance of this model is still significantly lower than the human ceiling (p=0.004). This shows that without task-specific training, but currently only with enormous model scale, acceptable LLM performance on our task is possible.

Comparing `BLOOM` and `Qwen2.5` models of a comparable size, we see that the more recent family, Qwen, achieves significantly better results in all cases, but only by a small margin (+4.2%, +6.1%, +5.1%, +4.9%, all with p<0.05). This result holds even though BLOOM was designed to have good performance on European languages. XLM-RoBERTa (large) is smaller by parameter count, but in Kodali et al.'s (2024) experiment, it performs best on their related CS task. Here, it performs better than others (performance significantly better or indistinguishable from all systems except the largest two `Llama-3` models), but it is inferior to `Llama-3.1 70B` and `405B`.

---
[7] There may be a difference in the number of tokens in the sentences in a minimal pair. It is possible that this problem is especially pronounced in our setup, as current LLM tokenizers tend to tokenize non-English text into more tokens (Petrov et al., 2023). Nevertheless, we do not normalise by number of tokens, as this has been shown to be an ineffective intervention (Ueda et al., 2024).

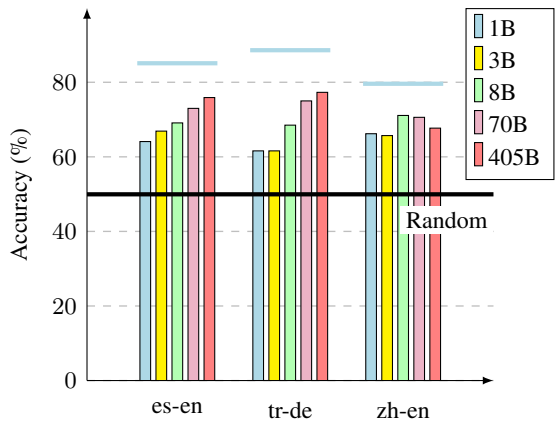

Figure 4: `Llama-3` model performance across three other language pairs. Horizontal light blue lines represent the accuracy achieved by our participants.

We perform an analysis of the extent to which the margin of the overall best LLM (`Llama-3.1 405B`) is correlated with human agreement on the task. Figure 3 shows our German–English samples separated into four categories, based on how many of the human participants agreed with the gold standard. Median LLM margin is significantly different between all pairs of categories (the same is true for mean margin, except between 0/3 and 1/3). Of those minimal pairs for which 3/3 bilinguals judged the observed CS sentence to be more acceptable, LLM probabilities between the first and third quartiles are all also higher for the observed CS sentence. Meanwhile, of those 22 minimal pairs for which all the bilinguals judged the manipulated variant as more acceptable, the LLM sided with the bilinguals (LLM probabilities between the first and third quartiles are higher for the variant). This is the first time that such similarity between LLMs and bilingual human CS preferences has been shown. We suspect that this is because previous experiments were either performed at much smaller scales, or not directly with LLM probabilities.

German and English are in the same language family and both are high-resource languages. Hence our results for this language pair may be less surprising than other possible language pairs. Results for Spanish–English, Turkish–German and Chinese–English are provided for the `Llama` models in Figure 4. The Spanish–English minimal pairs are derived from tweets. The numbers are near-identical to those from the German–English tweets, with consistent improvements for larger model size and the largest 405B model significantly outperforming all of its smaller counterparts.

The Turkish–German and Chinese–English data is derived from transcriptions of spoken CS. Figure 4 shows that Turkish–German trends are similar to before. This result is surprising, because prior work led us to expect that constraints in spoken CS follow different constraints to those in written CS (Gardner-Chloros and Weston, 2015) and all tested LLMs are stated to be trained on text only. Meanwhile, for Chinese–English, there is no improvement with model size (all comparisons have p>0.05). Poor performance for this language pair also holds for other LLMs, including the `Qwen2.5` LLM family (cf. Appx. Tab. 6). A contributing factor may be that these two languages are not written in the same script, because unlike in the case of other language pairs there are no tokens shared between the languages.

A similar Figure showing `Llama-3` model performance for all language pairs is provided in Appendix D. The `Llama-3.1 405B` model performs significantly better than all its smaller counterparts, except Danish–English (where it is statistically indistinguishable from the 3/8/70B models), Indonesian–English, Dutch–English, Swedish–English and Turkish–English (where it is indistinguishable from the 70B model).[8]

**POS-based analysis** We will now perform a more fine-grained analysis of the acceptability differential in the minimal pairs. We test the long-standing theoretical claim that words of particular grammatical functions tend to be more switchable (Joshi, 1982), in particular the extent to which the POS (which can represent a closed or open class of words) of the manipulated word in our minimal pairs affects the absolute LLM margin.[9] We now have the tools at hand to test this. We assume that larger LLM margin equates to a larger difference in acceptability in a minimal pair. This is supported by our results so far, which show that probabilities from this LLM align well with the gold standard and are correlated with human judgments. We now test, using our best LLM, whether larger absolute LLM margins are correlated with certain POS.

---

[8] Numerical results for all models and language pairs are provided in Appendix E.

[9] In this analysis, we include POS for which we have at least 10 minimal pairs. We obtain the POS of the changed word from its aligned word in the monolingual translation; we only include minimal pairs in this analysis where the changed word is aligned to a single identical word in the translation.

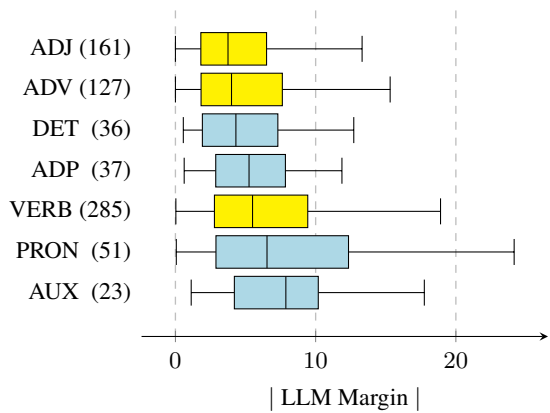

Figure 5: Absolute LLM margin vs. part of speech of the manipulated word (de-en). Blue represents closed-class items, yellow represents open-class items.

Results for German–English are given in Figure 5, colour-coded by closed and open-class items.[10] (Results of the same analysis for each of the other language pairs is shown in Appendix B.) Open class words effect an average LLM margin of 3.8 for adjectives (ADJ), 4.0 for adverbs (ADV) and 5.5 for verbs (VERB). Closed class words effect an LLM margin of 4.3 for determiners (DET), 5.3 for adpositions (ADP), 6.5 for pronouns (PRON) and 7.9 for auxiliaries (AUX). For this language pair we do not find a statistically significant overall difference (p=0.077). But overall we find consistent trends that changing closed-class words lead to a larger effect size than changing open-class words (significant for all language pairs except German–English and Italian–English). This is the first time that such direct and empirical testing of linguistic hypotheses about CS is possible. The methodology used in prior studies was able to show the significance of POS in CS corpora (Soto et al., 2018; Chi and Bell, 2024), but it cannot be used to directly test the claim above, because without manipulation of the CS and formation of a minimal pair, the switched variant would never be observed. Our benchmark allows for many kinds of hypothesis testing. In our companion paper (Sterner and Teufel, 2025), we test a more involved linguistic hypothesis concerning the connection between syntax and switch points in CS.

---

[10] It is worth noting that our sets of minimal pairs are not balanced across open vs. closed classes; there are fewer minimal pairs with changes to closed-class words (147 vs. 573 in this experiment). This is likely a result of our decision to enforce that every minimal pair we create for a particular language pair contains a lexical change not present in any other minimal pair.

In sum, our results show that judging the acceptability differential in our minimal pairs is a hard task. The patterns that determine what CS bilinguals generally judge as acceptable are almost certainly complex; this is in part demonstrated by the lack of one widely accepted theory of the phenomenon. That being said, we are the first to measure the degree to which LLMs can model CS acceptability.

## 4 Conclusion

Acceptability judgments and minimal pairs are two fundamental ideas in linguistic analysis. In this work, we have integrated these ideas in an automatic evaluation methodology for CS. The task is to determine which sentence in a minimal pair is the naturally occurring CS sentence. We present here the ACS benchmark, which contains up to 1,000 minimal pairs of CS each in 11 language pairs. Using human judgments, we demonstrated that there is an acceptability differential in the minimal pairs and that this is true across a diverse range of language pairs. The novelty of this approach is that any automatic system can now be tested on these minimal pairs; we no longer require ad-hoc human judgments of any particular system's output as was necessary in previous work. Experiments with multilingual LLMs show that only the largest current models are able to consistently make the distinction between sentences in the minimal pairs. This work makes it possible to track the progress of future LLMs in their ability to use CS in the same way as bilinguals.

## Limitations

We collect most of our CS data from Twitter. On the one hand, this data collection enables us to scale the methods so that we can generate enough minimal pairs that comply with our conditions. On the other hand, Twitter data is also noisy, and this noise can contribute to the acceptability differential. For example, we observed a German–English example where the incorrectly spelt English 'belive' was automatically translated into the correctly spelt German 'glauben'.

Errors from the automatic processes restrict the quality of the minimal pairs in our benchmark. We found some minimal pairs that did not contain any

CS at all, either because the minimal pair removed all CS or because the observed sentence did not exhibit CS (see e.g. Appx. Ex. 41). The quality of the benchmark could be improved by further filtering along these lines. Our processes also assume CS occurs between only two languages, which fails to support multilingual CS between many languages (see e.g. the Dutch–English–German sentence in Appx. Ex. 21a).

Also, using Twitter data does not enable us to collect socio-demographic profiles (such as age, location, language proficiency) of the authors of the tweets in our benchmark. We cannot, therefore, be certain of the linguistic variation of CS present in our benchmark, be it location-based, register-based or other social-demographics (Doğruöz et al., 2023). We cannot even be sure that every CS sentence we observe was written by a human, rather than by an automatic system. On top of that, we also also cannot be sure whether the human participants in our study are representative of the authors of the CS we use.

Our human study was, except for German–English, limited to one participant per language pair for a subset of our testing data. From a methodological viewpoint, a follow-up study with more participants and possibly more language pairs could be advantageous.

There is a question about whether any of the models we tested were trained on the data that makes up the benchmark. The best performing LLMs were from the `Llama-3` (MetaAI, 2024) family, and the pre-training data for these models is not publicly known. There are reasons to believe that the German–English test tweets were not included: the source corpus is, to the best of our knowledge, not part of any open-access corpus. Additionally, we used the most recent data that was available for every language, as this may lower the chance of our test data having appeared in the models' pre-training data.

## Ethics Statement

Our work involves working with and releasing a large corpus of social media posts. This raises privacy concerns. We do not collect any data about the authors of the posts. The majority of the data we release has been anonymised: for the German–English data, we replaced mentions with a "@USER" placeholder and URLs with "HTTPURL". DeLucia et al.'s corpus is already anonymized in this same way. The text from the two spoken corpora were left as is.

In our human experiment, we follow the ethics guidelines of the Cambridge University Department for Computer Science and Technology, asking for formal consent from our participants in the experiment, paying them for their time and informing them of their rights.

## Acknowledgements

The first author was supported by Cambridge University Press & Assessment and a Cambridge Trust studentship supported by Pembroke College, Cambridge. We thank Louis Cotgrove, Martin Durrell, Markus Frohman, Luna Hawkins and Edoardo Ponti for their comments.

# A Pre-Processing Details

**German–English Pre-Processing**   Our earlier German–English token-level language identification system (Sterner and Teufel, 2023) is inexpensive to run, because it is based on lookups in large wordlists. Our compilation of this pre-processed data uses the official implementation, with a number of adaptations for our purposes.

1. One of the problems in our previous corpus was Swiss German sentences, because we decided to include Swiss and Austrian wordlists in the compilation process. For the work here, we do not include these wordlists.
2. We also exclude tweets which contain words with an umlaut but not in the German wordlist. Such words are almost never German/English words.
3. We found many monolingual tweets which arise due to incorrectly identified interlingual homographs and other shared words between the languages. We search for German words in a bilingual dictionary (`dict.cc`) which have identical corresponding English entries, and vice versa. We remove all such words from their wordlists of each language. A second aspect of this problem is with interlingual homographs. These words cause a bias in their corpus towards tweets with a small number of incorrectly disambiguated interlingual homographs. We mark all interlingual homographs as unknown, which is the same category as words not in any of the wordlists or caught by any of their other steps.
4. We only keep tweets with fewer than 50% unknown words.

**Borrowed words**   Borrowed words are words of foreign origin used in one language that are fully assimilated and hence considered part of that language. They form a grey area in language change, and for our purposes should not be altered in a minimal pair.

We collect lists of borrowed words for each language pair, that is lists of words borrowed from each Lang1 to Lang2 and vise versa. If `AnE` tags a word as Lang1, and in fact the lists reveal it is borrowed from Lang2, we tag it as language neutral and vise versa. To gather the lists of borrowings, we use information from Wiktionary. Table 3 shows how many we find in the language pairs of interest and a few samples for each. In particular, we use the machine-parsed version of Wiktionary from Ylonen (2022).

**Named Entities**   `AnE` identifies named entities, and we further identify named entities in our monolingual translations, but we found many are still missed. We mark consecutive words containing over 75% capitalized words as named entities.

**Chinese–English Pre-Processing**   The regular expression `[\u4e00-\u9fff]` distinguishes Chinese characters from all other characters. The text is tokenized by passing the Chinese and English segments to neural Chinese/English tokenizers separately and concatenating the resulting words.

| Lang | # | to English | # | from English |
|------|------|------|------|------|
| da | 153 | leverpostej, troll, skol | 335 | whistleblower, bodybuilder, clean |
| de | 3891 | Trautwein, schnitzel, müsli | 1307 | downloaden, Limerick, happy |
| es | 3579 | guacamole, piña colada, enchilada | 1252 | blockchain, film, bróker |
| fr | 6254 | triolet, moulinette, arpent | 1013 | woofing, nétiquette, ragequitter |
| it | 3614 | beccafico, DiFiore, piadine | 891 | George, Utah, mister |
| nl | 762 | Hoppes, Hoefs, Barten | 1319 | falsificationisme, nonsens, webcam |
| sv | 355 | ombudsman, smorgasbord, fika | 566 | shopping, coach, internet |
| tr | 171 | doner kebab, Türk, Muş | 207 | kok, server, sensör |

Table 3: Borrowings indexed in the English Wiktionary. # refers to the total number of borrowings in that language pair. Each borrowing is hyperlinked to its Wiktionary page.

## B Results of POS-Based Analysis

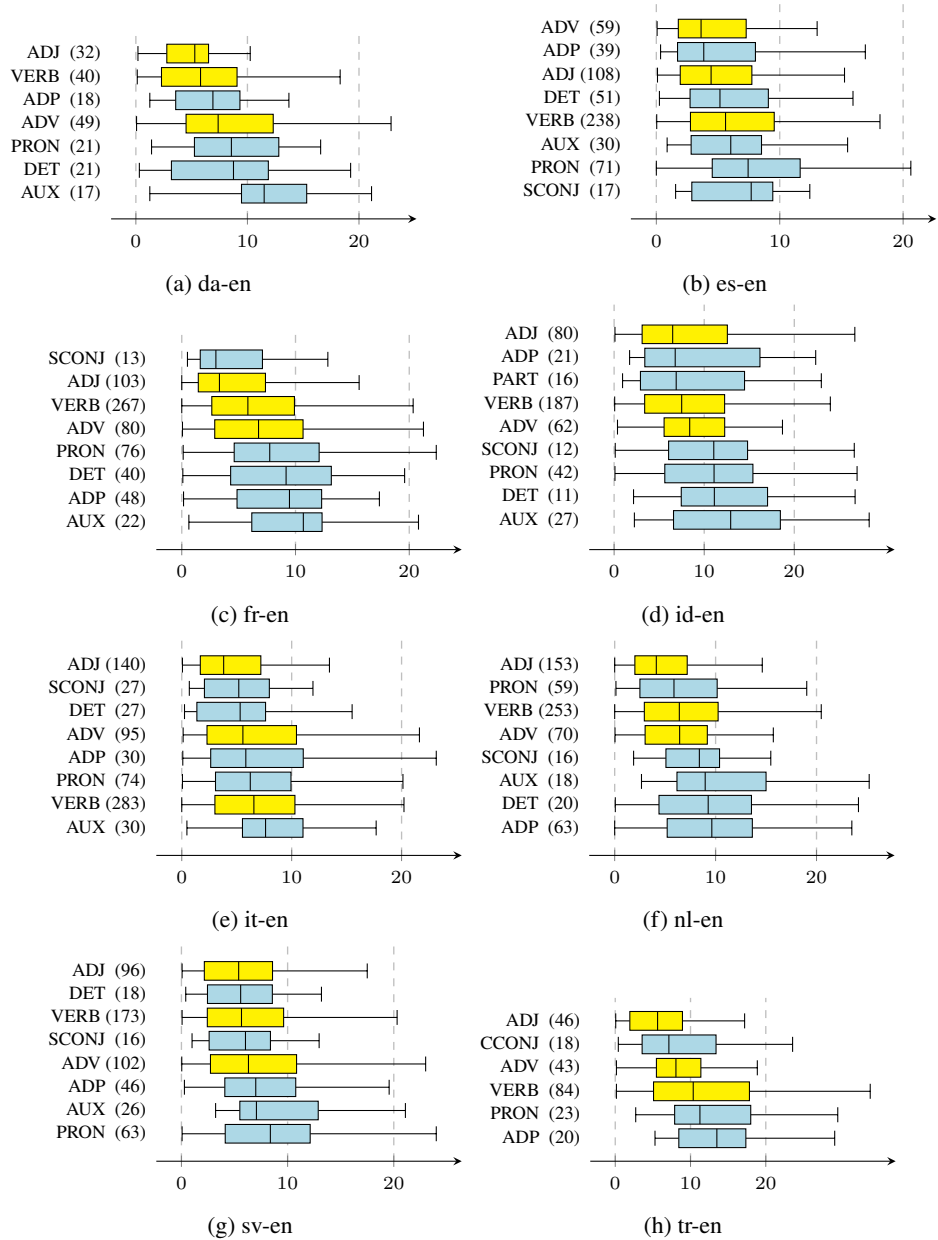

Figure 6: Absolute LLM margin vs. part of speech of minimal pair change for tweet-derived minimal pairs. Blue represents closed-class items, yellow represents open-class items.

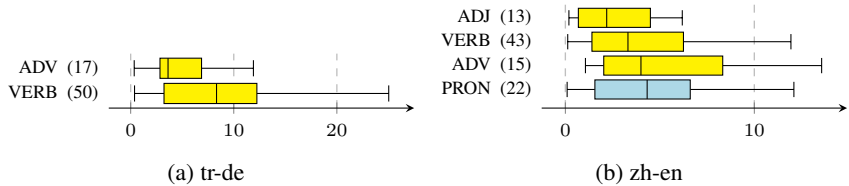

Figure 7: Absolute LLM margin vs. part of speech of minimal pair change for data derived from transcriptions of spoken CS. Blue represents closed-class items, yellow represents open-class items.

## C Benchmark

Table 4 gives statistics of the ACS benchmark and the data used for its construction. The reported number of CS tweets and CS sents refers to text with at least one switch point and at least two neighbouring tokens of each language. The number of minimal pairs (MPs) is after all the further constraints we apply for a sentence to quality to become a minimal pair (we applied a cap of 1,000).

The minimal pairs we release are planned for testing. For German–English, our processing pipeline is robust and we have a large corpus of tweets to start from. Hence, we also release validation and training data for this language pair, although the evaluation framework we use here does not require it. The German–English training data is from 04/2019–12/2021 and validation data is from 01/2022–06/2022. The German–English testing data is more recent (from tweets in 07/2022–02/2023).

| Codes | Language Pair | # input tweets | # CS tweets | # CS sents | # MPs | # judgments | Split |
|---|---|---|---|---|---|---|---|
| de-en | German-English | 27,474,736 | 130,619 | 113,789 | 1,000 | 335 x 3 | test |
| | | 21,434,579 | 147,513 | 127,601 | 1,932 | - | validation |
| | | 100,285,504 | 840,946 | 713,605 | 7,501 | - | train |
| es-en | Spanish-English | 10,000,000 | 149,133 | 81,582 | 860 | 201 x 1 | test |
| da-en | Danish-English | 1,867,709 | 28,689 | 17,627 | 330 | 201 x 1 | test |
| fr-en | French-English | 10,000,000 | 348,856 | 234,520 | 1,000 | 201 x 1 | test |
| id-en | Indonesian-English | 10,000,000 | 826,865 | 614,049 | 1,000 | 201 x 1 | test |
| it-en | Italian-English | 10,000,000 | 273,651 | 161,489 | 1,000 | 201 x 1 | test |
| nl-en | Dutch-English | 10,000,000 | 390,189 | 237,692 | 1,000 | 201 x 1 | test |
| sv-en | Swedish-English | 3,263,865 | 72,618 | 46,286 | 760 | 201 x 1 | test |
| tr-en | Turkish-English | 10,000,000 | 149,526 | 88,672 | 553 | 201 x 1 | test |
| tr-de | Turkish–German | - | - | - | 216 | 201 x 1 | test |
| zh-en | Chinese–English | - | - | - | 201 | 201 x 1 | test |

Table 4: Summary of the ACS corpus and benchmark.

### C.1 Glosses and Translations

(4) a. *@USER **And I said maybe** <u>etwas</u> leiser singer, sonst ruf ich die Polizei*
    b. *@USER **And I said maybe** <u>a little</u> leiser singen, sonst ruf ich die Polizei*
       quieter sing, otherwise call I the police

   '@USER And I said maybe sing a little quieter, or I'll call the police'

(5) a. ***Would do it** <u>myself</u> om inte make var bilmek och nördig med det.*
    b. ***Would do it** <u>själv</u> om inte make var bilmek och nördig med det.*
       if not husband was mechanic and nerdy about it

   'Would do it myself if my husband was not a mechanic and into that stuff'

(6) a. 同学们 会 大概 <u>知道</u> ***what they want to do in the future***
    b. 同学们 会 大概 <u>know</u> ***what they want to do in the future***
       Students will generally

   'Students generally know what they want to do in the future'

## C.2 Danish–English

(7) a. **i am** <u>begging</u> en eller anden få et crush på mig
    b. **i am** <u>tigger</u> en eller anden få et crush på mig

(8) a. **the way** jeg var <u>mere</u> **excited over at** være på skolen hvor de filmede pagten end rent […]
    b. **the way** jeg var <u>**more**</u> **excited over** at være på skolen hvor de filmede pagten end rent […]

(9) a. Men han har også <u>**a way with the ladies**</u>.
    b. Men han har også <u>en</u> **way with the ladies**.

## C.3 Spanish–English

(10) a. si mi prima también <u>**walk**</u> **around central park**
     b. si mi prima también <u>camina</u> **around central park**

(11) a. **I pranked myself** <u>sintiéndome</u> culpable por ti, cuando realmente eras tu el cdv
     b. **I pranked myself** <u>feeling</u> culpable por ti, cuando realmente eras tu el cdv

(12) a. @USER Amor, **talk to** <u>him</u> y establece acuerdos: ((
     b. @USER Amor, **talk to** <u>él</u> y establece acuerdos: ((

## C.4 German–English

(13) a. müsste vorher noch bissl wachsen lassen wegen **undercut but i wanna try it**
     b. müsste vorher noch bissl wachsen lassen <u>**because of**</u> **undercut but i wanna try it**

(14) a. wenn jmd beleidigt dann <u>direkt</u> **no friendship**, **thats a obvious for me**
     b. wenn jmd beleidigt dann <u>**directly**</u> **no friendship**, **thats a obvious for me**

(15) a. **at this point I'm convinced** <u>here</u> gibts nur weirdos
     b. **at this point I'm convinced** <u>hier</u> gibts nur weirdos

## C.5 Italian–English

(16) a. e **tbh I would be pissed** <u>anche</u> se uno fosse più bravo di me in qualcosa che faccio da […]
     b. e **tbh I would be pissed** <u>too</u> se uno fosse più bravo di me in qualcosa che faccio da […]

(17) a. […]sola in un fosso smatto veramente e Non Ironicamente mi <u>**do upset in the worst way**</u>
     b. […]sola in un fosso smatto veramente e Non Ironicamente mi <u>rende</u> **upset in the worst way**

(18) a. […]che non ti caga proprio perché tu non vedresti l' ora di litigarci per <u>**far**</u> **show** #gfvip
     b. […]che non ti caga proprio perché tu non vedresti l' ora di litigarci per <u>farti vedere</u> **show** #gfvip

## C.6 Dutch–English

(19) a. Wat mooi: een héél concreet voorbeeld van **exploring the potential of nature to improve** […]
     b. Wat mooi: een héél concreet voorbeeld van het verkennen **the potential of nature to improve** […]

(20) a. lotte **is** shook omdat ik nog nooit **divergent** gezien heb lol
     b. lotte **is** erg geschokt omdat ik nog nooit divergent gezien heb lol

(21) a. Maar echt, als dit serieus is, **go see** ein Herr Doctor!
     b. Maar echt, als dit serieus is, **go see** a Herr Doctor!

## C.7 Swedish–English

(22) a. att han **pops up in my mind randomly** skrämmer mig, sluta ockupera mina tankar tack
     b. att han dyker **up in my mind randomly** skrämmer mig, sluta ockupera mina tankar tack

(23) a. Mindre pladder och mer **implicit story** hade passat mig bättre - det är inte min favoritgenre.
     b. Mindre pladder och **more implicit story** hade passat mig bättre - det är inte min favoritgenre.

(24) a. **Would do it myself** om inte make var bilmek och nördig med det.
     b. **Would do it** själv om inte make var bilmek och nördig med det.

## C.8 French–English

(25) a. **on my road to** la deuxième dose alors que je suis déjà épuisée
     b. **on my road** vers la deuxième dose alors que je suis déjà épuisée

(26) a. **Watch us** revenir comme des queens dans notre château
     b. **Watch us come back** comme des queens dans notre château

(27) a. **still accurate** mais je me contrôle mieux
     b. **still** précis mais je me contrôle mieux

## C.9 Indonesian–English

(28) a. **I shopping** luar sbb nk compare dgn harga dlm shopee
     b. **I** belanja luar sbb nk compare dgn harga dlm shopee

(29) a. @USER @USER Bismillah, semoga mama sponsor **get fast recover** dan kembali sehat, aamiin
     b. @USER @USER Bismillah, semoga mama sponsor **get fast** sembuh dan kembali sehat, aamiin

(30) a. Sebab nya, asal aku lapor **or craving something** mesti **notification** dia **pop up**!
     b. Sebab nya, asal aku report **or craving something** mesti **notification** dia **pop up**!

## C.10 Turkish–English

(31) a. ben de **feel the festival getting closer** ama katılamiyo olmak cigerimi yakiyo
 b. ben de hissediyorum **the festival getting closer** ama katılamiyo olmak cigerimi yakiyo

(32) a. Her zaman derim **better deserves better**
 b. Her zaman derim daha iyi **deserves better**

(33) a. […]bu kadar insanın **willingly** kendini kandırma sebebi **feels good and keeps you sane** olması da
 b. […]bu kadar insanın **willingly** kendini kandırma sebebi **feels good and keeps you** akıllı olması da

## C.11 Turkish–German

(34) a. **Er ist gestorben** iki üç ay önce öldü.
 b. **Er ist gestorben** zwei ay önce öldü.

(35) a. İngilizce öyle fazla da konuşmadım **irgendwie in dem Bachelor und ich habe eben das Gefühl** sanki böyle unuttum gibime geliyor.
 b. İngilizce öyle fazla da konuşmadım bir şekilde **in dem Bachelor und ich habe eben das Gefühl** sanki böyle unuttum gibime geliyor.

(36) a. **In dem Winter ist eben so** bir kazağı üç kere giyinebiliyorsun **bevor du es waschen musst**.
 b. **In dem Winter ist eben so bir** kazağı üç kere anziehen **bevor du es waschen musst**.

## C.12 Chinese–English

(37) a. **i think** 他们在当时 **shouldn't have made empty promises to** 安抚我的心
 b. **i think** 他们在当时 **shouldn't have made empty promises to** reassure 我的心

(38) a. 本来还以为 **i was getting used to it** until 我在考试期间搬回家的时候 **found that my sleep** […]
 b. 本来还以为 **i was getting used to it** 直到我在考试期间搬回家的时候 **found that my sleep** […]

(39) a. 如果你就比如说呃做些好事啊或者说你及时的 **pay in** 你的你的 **tax** 啊或者什么的你的 **credit rate** 会增加
 b. 如果你就比如说呃做些好事啊或者说你及时的缴纳 **in** 你的你的 **tax** 啊或者什么的你的 **credit rate** 会增加

## C.13 German–English Disagreement with the Gold Standard

In the following minimal pairs all three human participants selected sentence b, preferring the manipulated sentence over the observed CS sentence.

(40) a. Nein, so einfach lasse ich die Politischen **Powers that were** da nicht entkommen.
    b. Nein, so einfach lasse ich die Politischen **Powers that were** there nicht entkommen.

(41) a. Ihr linken werdet auf dem Altar der **wokeness** due nächsten Wahlen opfern.
    b. Ihr linken werdet auf dem Altar der **wokeness** die nächsten Wahlen opfern.

(42) a. […]und dann suddenly **is** februar **and i still haven't planed anything**
    b. […]und dann plötzlich **is** februar **and i still haven't planed anything**

(43) a. **I mean** better als alle anderen Optionen
    b. **I mean** besser als alle anderen Optionen

(44) a. **Ranking the people** im Wartezimmer von 1-10 **on how** wahrscheinlich es wäre für **me to win in a fight**.
    b. **Ranking the people** im Wartezimmer von 1-10 **on how** likely es wäre für **me to win in a fight**.

(45) a. 'Leider' hat der von Mir gewuenschte Artist, der auch in der Vergangenheit Coverart fuer mich gemacht hat, seine **Commissions** closed da er vom **Freelancer** zum **Full Employment** aufgestiegen **is**
    b. 'Leider' hat der von Mir gewuenschte Artist, der auch in der Vergangenheit Coverart fuer mich gemacht hat, seine **Commissions** geschlossen da er vom **Freelancer** zum **Full Employment** aufgestiegen **is**

(46) a. Das macht mich wütend **and I dunno why**
    b. Das macht mich **angry and I dunno why**

(47) a. **I** be kurz vor **a relationship then get an** ick sodass Icj die Person nicht mehr ansehen kann
    b. **I** bin kurz vor **a relationship then get an** ick sodass Icj die Person nicht mehr ansehen kann

(48) a. ist bisschen unnötig **but I enjoy drama** so.
    b. ist bisschen **unnecessary but I enjoy drama** so.

(49) a. **I would cheat** sofort für und mit euch.
    b. **I would cheat** immediately für und mit euch.

(50) a. **I am once again asking** inwiefern das Konzept FLINTA irgendeinen Sinn ergibt.
    b. **I am once again asking** how das Konzept FLINTA irgendeinen Sinn ergibt.

(51) a. @USER @USER find das jetzt aber not so **heavy** und **it helps a lot with** Hautbild
    b. @USER @USER find das jetzt aber nicht so **heavy** und **it helps a lot with** Hautbild

(52) a. **I get the** Innenleben und du den **crunchy part**
    b. **I get the** Innenleben und du den knusprigen **part**

(53) a. Daraus **it's** <u>wahrscheinlich</u> eine große Skepsis gegenüber Normen die im westlichen Länder formuliert würden.
  b. Daraus **it's** <u>probably</u> eine große Skepsis gegenüber Normen die im westlichen Länder formuliert würden.

(54) a. **But not because of the** schwachsinnigen "menschgemachten" Klimawandel **and also not because of the** bescheuerte angebliche Pandemie, **but because of the really very** <u>gefährliche</u> sogenannte 'Impfung'!!!
  b. **But not because of the** schwachsinnigen "menschgemachten" Klimawandel **and also not because of the** bescheuerte angebliche Pandemie, **but because of the really very** <u>dangerous</u> sogenannte 'Impfung'!!!

(55) a. **i miss you** <u>just</u> und ich hab mich gerade in den nächsten jahren in der schule […]
  b. **i miss you** <u>gerade</u> und ich hab mich gerade in den nächsten jahren in der schule […]

(56) a. (und hab <u>probably</u> adhd/autism **but thats not the point**)
  b. (und hab <u>wahrscheinlich</u> adhd/autism **but thats not the point**)

(57) a. […]**also i believe i can** <u>fahr</u> sozusagen #haltdiefressejudithholofernes
  b. […]**also i believe i can** <u>drive</u> sozusagen #haltdiefressejudithholofernes

(58) a. **because of** Hallux Valgus OP **and** irgendwas **with** Schräubchen **and** Plättchen **and** Arthrose **and a little bit of** <u>komplizierte</u> Geschichte
  b. **because of** Hallux Valgus OP **and** irgendwas **with** Schräubchen **and** Plättchen **and** Arthrose **and a little bit of** <u>complicated</u> Geschichte

(59) a. @USER @USER **I wanna see the** <u>gekauft</u> Zeichen
  b. @USER @USER **I wanna see the** <u>purchased</u> Zeichen

(60) a. wenigstens weiß ich jetzt, dass mein **eyeliner** <u>actually</u> wasserfest ist **i guess**
  b. wenigstens weiß ich jetzt, dass mein **eyeliner** <u>tatsächlich</u> wasserfest ist **i guess**

(61) a. **i spoke to my dad and he said he might** <u>überweisen</u> meine schwester **sum for me so she can gimme it**
  b. **i spoke to my dad and he said he might** <u>transfer</u> meine schwester **sum for me so she can gimme it**

## D  `Llama-3` Model Performance

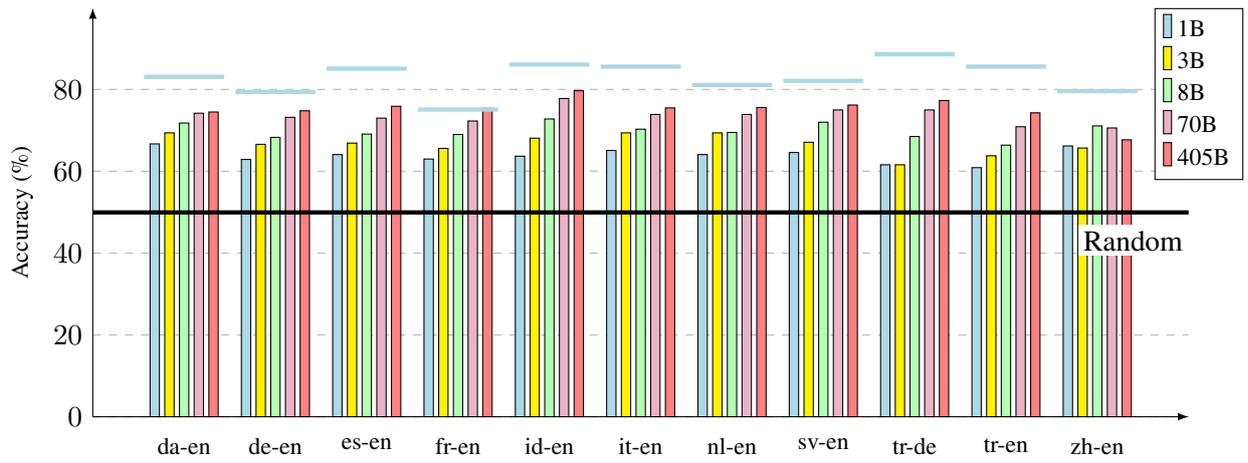

Figure 8: `Llama-3` model performance. Horizontal light blue lines represent the accuracy achieved by our participants.

# E   Numerical Results

|       | 0.6M | 1.1B | 1.7B | 3B   | 7.1B | 176B |
|-------|------|------|------|------|------|------|
| da-en | 54.2 | 58.2 | 57.3 | 62.7 | 60.9 | 67.3 |
| de-en | 54.6 | 54.4 | 57.5 | 58.8 | 61.2 | 64.2 |
| es-en | 61.2 | 59.4 | 65.5 | 65.9 | 68.3 | 72.3 |
| fr-en | 65.1 | 65.6 | 67.8 | 68.7 | 71.0 | 72.6 |
| id-en | 63.7 | 64.3 | 64.8 | 66.7 | 69.3 | 75.1 |
| it-en | 56.0 | 55.3 | 57.3 | 60.4 | 61.1 | 64.7 |
| nl-en | 57.2 | 59.6 | 59.4 | 59.4 | 62.0 | 65.1 |
| sv-en | 60.3 | 61.7 | 62.5 | 62.4 | 62.8 | 64.6 |
| tr-de | 54.6 | 53.7 | 56.9 | 58.8 | 57.9 | 60.6 |
| tr-en | 50.5 | 53.9 | 54.6 | 55.7 | 56.1 | 58.8 |
| zh-en | 59.7 | 67.2 | 62.7 | 65.7 | 73.1 | 67.2 |

Table 5: Numerical results for the BLOOM LLM family.

|       | 0.5B | 1.5B | 3B   | 7B   | 14B  | 32B  | 72B  |
|-------|------|------|------|------|------|------|------|
| da-en | 64.5 | 66.7 | 66.4 | 68.5 | 71.8 | 70.6 | 69.7 |
| de-en | 58.8 | 63.6 | 63.9 | 66.1 | 66.6 | 66.0 | 67.4 |
| es-en | 59.8 | 64.3 | 65.2 | 66.2 | 68.6 | 70.6 | 71.0 |
| fr-en | 62.0 | 64.3 | 66.9 | 68.6 | 71.1 | 71.2 | 71.9 |
| id-en | 61.2 | 64.2 | 68.1 | 70.4 | 71.7 | 70.4 | 74.6 |
| it-en | 60.0 | 64.5 | 64.8 | 68.3 | 68.6 | 71.4 | 72.0 |
| nl-en | 62.0 | 64.8 | 66.2 | 68.5 | 69.0 | 68.8 | 70.5 |
| sv-en | 63.7 | 64.1 | 65.1 | 66.1 | 69.3 | 68.6 | 71.4 |
| tr-de | 60.6 | 63.0 | 63.4 | 66.7 | 72.2 | 68.1 | 70.8 |
| tr-en | 56.2 | 60.2 | 60.9 | 62.4 | 64.0 | 64.6 | 69.1 |
| zh-en | 64.7 | 67.2 | 67.7 | 70.1 | 66.7 | 70.6 | 71.1 |

Table 6: Numerical results for the Qwen2.5 LLM family.

|       | 1B   | 3B   | 8B   | 70B  | 405B |
|-------|------|------|------|------|------|
| da-en | 66.7 | 69.4 | 71.8 | 74.2 | 74.5 |
| de-en | 62.9 | 66.6 | 68.3 | 73.2 | 74.8 |
| es-en | 64.1 | 66.9 | 69.1 | 73.0 | 75.9 |
| fr-en | 63.0 | 65.6 | 69.0 | 72.3 | 75.4 |
| id-en | 63.7 | 68.1 | 72.8 | 77.8 | 79.7 |
| it-en | 65.1 | 69.4 | 70.3 | 73.9 | 75.5 |
| nl-en | 64.1 | 69.4 | 69.5 | 73.9 | 75.6 |
| sv-en | 64.6 | 67.1 | 72.0 | 75.0 | 76.2 |
| tr-de | 61.6 | 61.6 | 68.5 | 75.0 | 77.3 |
| tr-en | 60.9 | 63.8 | 66.4 | 70.9 | 74.3 |
| zh-en | 66.2 | 65.7 | 71.1 | 70.6 | 67.7 |

Table 7: Numerical results for the Llama-3 LLM family.

|       | 7B   | 13B  |
|-------|------|------|
| da-en | 68.2 | 70.6 |
| de-en | 66.3 | 67.2 |
| es-en | 63.7 | 66.2 |
| fr-en | 66.3 | 66.6 |
| id-en | 69.5 | 70.1 |
| it-en | 66.2 | 64.9 |
| nl-en | 65.1 | 65.8 |
| sv-en | 67.1 | 66.8 |
| tr-de | 65.3 | 72.2 |
| tr-en | 59.7 | 61.3 |
| zh-en | 68.7 | 65.2 |

Table 8: Numerical results for the `OLMo` LLM family.

|       | 1.7B | 9B   |
|-------|------|------|
| da-en | 69.1 | 72.4 |
| de-en | 62.4 | 68.6 |
| es-en | 62.9 | 65.7 |
| fr-en | 64.5 | 68.1 |
| id-en | 57.7 | 59.4 |
| it-en | 64.6 | 70.4 |
| nl-en | 66.2 | 71.4 |
| sv-en | 66.4 | 69.2 |
| tr-de | 64.4 | 70.8 |
| tr-en | 60.4 | 67.6 |
| zh-en | 64.2 | 63.2 |

Table 9: Numerical results for the `EuroLLM` family.

|       | base | large |
|-------|------|-------|
| da-en | 69.7 | 73.6  |
| de-en | 63.7 | 68.5  |
| es-en | 60.8 | 61.7  |
| fr-en | 60.8 | 64.5  |
| id-en | 71.4 | 75.6  |
| it-en | 61.5 | 67.0  |
| nl-en | 66.1 | 70.4  |
| sv-en | 70.1 | 73.6  |
| tr-de | 63.9 | 66.2  |
| tr-en | 60.4 | 64.9  |
| zh-en | 61.7 | 59.7  |

Table 10: Numerical results for the `XLM-RoBERTa` LLM family.